\documentclass[letterpaper]{article} %
\usepackage{aaai24}  %
\usepackage{times}  %
\usepackage{arydshln}
\usepackage{helvet}  %
\usepackage{courier}  %
\usepackage[hyphens]{url}  %
\usepackage{graphicx} %
\usepackage{amsmath}
\usepackage{natbib}  %
\usepackage{caption} %
\usepackage{algorithm}
\usepackage{algorithmic}
\usepackage{pifont}
\usepackage{amssymb}
\usepackage{newfloat}
\usepackage{listings}
\usepackage{multirow}
\usepackage{booktabs}
\usepackage{times}
\usepackage[inline]{enumitem}
\usepackage{latexsym}
\usepackage[T1]{fontenc}    
\usepackage[utf8]{inputenc}
\usepackage[inline]{enumitem}
\usepackage[dvipsnames]{colortbl}
\usepackage[utf8]{inputenc}
\usepackage[T1]{fontenc}
\usepackage{babel} 
\usepackage[toc,page]{appendix}
\usepackage{makecell}
\usepackage{soul}
\usepackage[dvipsnames, svgnames, x11names]{xcolor} 
\usepackage{verbatim}

\usepackage{xspace}
\usepackage{svg}
\definecolor{paleyellow}{HTML}{FFEF77}
\definecolor{paleorange}{HTML}{FBB068}
\definecolor{paleblue}{HTML}{65B2FF}
\definecolor{lemon}{HTML}{FDFFCC}

\definecolor{Gainsboro}{rgb}{0.86, 0.86, 0.86}

\definecolor{Gray}{gray}{0.95}
\definecolor{LightCyan}{rgb}{0.88,1,1}

\newcommand{\code}[1]{{\ttfamily#1}}

\newcommand{\first}[1]{\textcolor{green}{\textbf{#1}}}
\newcommand{\second}[1]{\textcolor{blue}{\textbf{#1}}}
\newcommand{\third}[1]{\textcolor{cyan}{\textbf{#1}}}
\newcommand{\fouth}[1]{\textcolor{teal}{\textbf{#1}}}

\DeclareCaptionStyle{ruled}{labelfont=normalfont,labelsep=colon,strut=off} %
\floatstyle{ruled}
\newfloat{listing}{tb}{lst}{}
\floatname{listing}{Listing}
\title{\textit{\ours}:
Iterative Tool Learning from Introspection Feedback \\ by Easy-to-Difficult Curriculum
 }

\author {
 Shen Gao\textsuperscript{\rm 1}\thanks{ Equal contribution.}, Zhengliang Shi\textsuperscript{\rm 1}\footnotemark[1], Minghang Zhu\textsuperscript{\rm 1}, Bowen Fang\textsuperscript{\rm 1}, \\
 Xin Xin\textsuperscript{\rm 1}, Pengjie Ren\textsuperscript{\rm 1}, Zhumin Chen\textsuperscript{\rm 1}, Jun Ma\textsuperscript{\rm 1}, Zhaochun Ren\textsuperscript{\rm 2}\thanks{ Corresponding author.}
}
\affiliations {
\textsuperscript{\rm 1}Shandong University, Qingdao, China\\
\textsuperscript{\rm 2}Leiden University, Leiden, The Netherlands \\
\{shengao, xinxin, renpengjie, chenzhumin, majun\}@sdu.edu.cn\\
\{shizhl, mhzhu\}@mail.sdu.edu.cn~~bwn.fang@gmail.com~~z.ren@liacs.leidenuniv.nl
}

\usepackage{bibentry}

\newcommand{\ours}{\textit{Confucius}\xspace}
\newcommand{\abb}{{ISIF}\xspace}
\newcommand{\full}{\textbf{I}terative \textbf{S}elf-instruct from \textbf{I}ntrospective \textbf{F}eedback\xspace}

\newcommand{\ie}{\emph{i.e.,}\xspace}
\newcommand{\aka}{\emph{a.k.a.,}\xspace}
\newcommand{\eg}{\emph{e.g.,}\xspace}

\begin{document}

\maketitle

\begin{abstract}

Augmenting large language models (LLMs) with external tools has emerged as a promising approach to extending the capability of LLMs.
Although some works employ open-source LLMs for the tool learning task, most of them are trained in a controlled environment in which LLMs only learn to execute the human-provided tools.
However, selecting proper tools from the large toolset is also a crucial ability for the tool learning model to be applied in real-world applications.
Existing methods usually directly employ self-instruction methods to train the model, which ignores differences in tool complexity.
In this paper, we propose the \textbf{\ours}, a novel tool learning framework to train LLM to use complicated tools in real-world scenarios, which contains two main phases:
(1) We first propose a multi-stage learning method to teach the LLM to use various tools from an easy-to-difficult curriculum;
(2) thenceforth, we propose the \full (\abb) to dynamically construct the dataset to improve the ability to use the complicated tool.
Extensive experiments conducted on both controlled and real-world settings demonstrate the superiority of our tool learning framework in real-world application scenarios compared to both tuning-free (\eg ChatGPT, Claude) and tuning-based baselines (\eg GPT4Tools).\footnote{The code is available at~\url{https://github.com/shizhl/CTL}}
\end{abstract}

\section{Introduction}\label{sec:introduction}

\begin{table*}[htbp]
\small
\centering

\setlength{\tabcolsep}{2mm}{
\begin{tabular}{@{}lccc cccc@{}}
\toprule
\textbf{Method}  
& \begin{tabular}[c]{@{}c@{}} \textbf{Base} \\ \textbf{Model} \end{tabular}    
&\begin{tabular}[c]{@{}c@{}} \textbf{Dataset} \\ \textbf{Construction}\end{tabular}    
& \begin{tabular}[c]{@{}c@{}}\textbf{Compositional} \\ \textbf{Reasoning} \end{tabular}  
& \begin{tabular}[c]{@{}c@{}}\textbf{Unseen Tool}\\ \textbf{ in Evaluation} \end{tabular}    
& \begin{tabular}[c]{@{}c@{}}\textbf{Candidates} \\ \textbf{Construction} \end{tabular}   
\\
\cmidrule(r){1-1}
\cmidrule(r){2-2}
\cmidrule(r){3-3}
\cmidrule(r){4-4}
\cmidrule(r){5-5}
\cmidrule(r){6-6}
\cmidrule(r){7-7}

Chameleon~\citep{chameleon}
&      GPT-4      
&         -         
&    \ding{52}      
& \ding{52}     
& Manually
\\ 

MMREACT~\citep{yang2023mm}
&    GPT-3.5   
&   - 
& \ding{52}
& \ding{52}     
&  Manually
\\ 

\specialrule{0em}{1pt}{1pt}
\cdashline{1-7}[6pt/6pt]
\specialrule{0em}{1pt}{1pt}

Toolformer~\citep{toolformer} 
&     GPT-J-6B  
&    In-Context Learning 
& \textcolor{red}{\ding{56} }        
& \textcolor{red}{\ding{56} }       
&  Manually
\\

GPT4Tools~\citep{gpt4tools}
&   Vicuna-13B  
&   Manually 
& \textcolor{red}{\ding{56} }    
&  \ding{52} 
&          Manually
\\

ToolAlpaca~\citep{toolalpaca}
&   Vicuna-13B    
&   Simulation 
& \textcolor{red}{\ding{56} }    
& \ding{52}       
&          Manually    
\\

APIBench~\citep{toolalpaca}
&   LLaMA-7B    
&   Manually 
&  \textcolor{red}{\ding{56} }  
 &  \textcolor{red}{\ding{56} }   
&          Manually     
\\

ToolBench~\citep{xu2023tool}
&    LLaMA-30B     
&     Self-Instruct  
&  \textcolor{red}{\ding{56} }   
 &  \textcolor{red}{\ding{56} }   
& Manually
\\ 

\textbf{Ours}
&    LLaMA-7B      
&   {\abb}     
&   \ding{52} 
&     \ding{52} 
&  Retrieval
\\ 

\bottomrule
\end{tabular}

}

\caption{Comparison of related works. 
\textbf{Dataset construction} denotes the method of obtaining the training dataset. 
\textbf{Compositional reasoning} indicates whether compositional reasoning is required when answering the user query.
\textbf{Candidates construction} indicates whether the candidate set is carefully constructed manually, or obtained through the same retrieval method as the real application scenario.
}
\label{tab:comparison}
\end{table*}

The task of tool learning aims to unleash the power of large language models (LLMs) to effectively interact with various tools to accomplish complex tasks~\citep{toolw}. 
By integrating LLM with APIs, we can greatly expand their utility and empower LLM to serve as an efficient intermediary between users and the vast ecosystem of applications~\citep{webcpm,Jin2023GeneGPTAL,Park2023GenerativeAI}.
Existing tool learning approaches can be divided into two categories: tuning-free and tuning-based methods.
The former ones leverage the proprietary LLMs, such as ChatGPT or GPT-4, to interact with various tools to solve complex tasks.
These methods prompt the proprietary LLMs with demonstrations of tool usage.
However, the only way for the proprietary LLMs to access the user-defined tools is the prompt~\citep{api-bank}.
Thus, the limited context length of LLMs restricts the application of massive tools.
In contrast, the tuning-based methods fine-tune open-source LLMs to memorize and understand external tools by explicitly training on elaborate datasets~\citep{api-bank,toolformer}.
The majority of these methods~\cite{toolw} first use Self-Instruct technique to collect tool-use data from proprietary LLMs and then fine-tune an open-source model.
Since the training data only contains a limited range of tools, most turn-based methods lack the capability to generalize to unseen tools (tools outside the training data).
In Table~\ref{tab:comparison}, we list several cutting-edge tool-use LLMs.

\begin{figure}[t]
        \centering
	\includegraphics[width=0.47\textwidth]{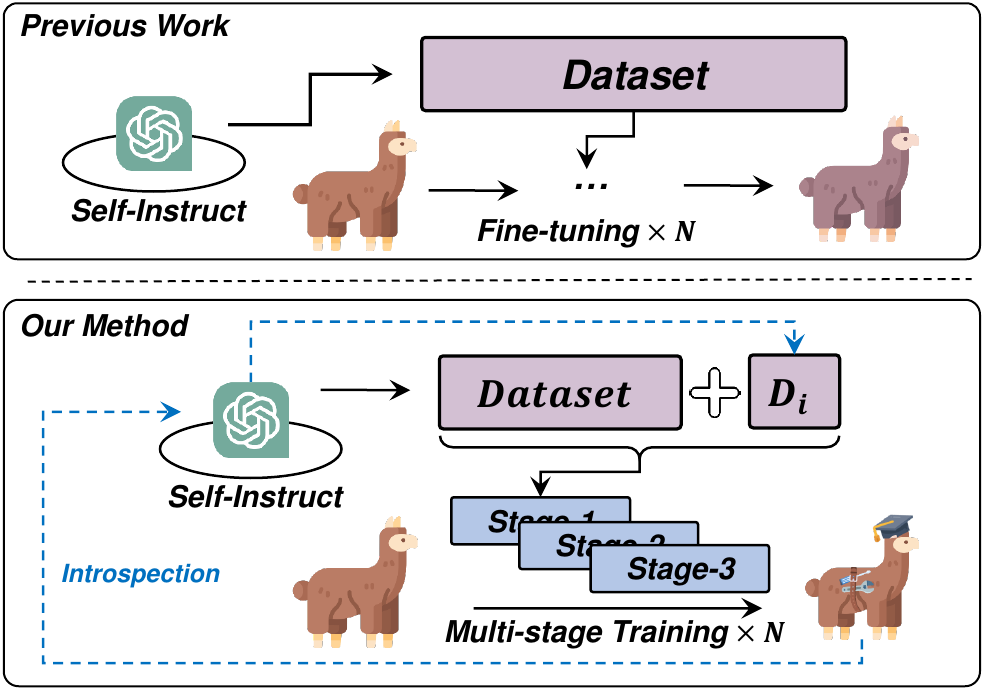}
        \caption{
       Comparison between the existing tuning-based tool learning methods and our \ours.
       Instead of using a pre-constructed dataset, we propose an iterative data construction framework with multi-stage learning to train the tool-use model effectively.
       }
 \label{fig:intro}
\end{figure}

As shown in Figure~\ref{fig:intro}, most existing methods directly provide a minimal essential toolset to LLMs without redundant tools.
However, when adapting to real-world applications,  LLMs typically face a large toolset that contains various tools across different tasks.
Thus, how to teach LLMs to select an appropriate tool from the candidates becomes the \textbf{first challenge}.

Intuitively, the difficulty of using different tools is not the same. Some tools are used in different ways in different scenarios, so more attention should be paid to using such complicated tools during model training. 
For example, the Google Map tool for exploring the surrounding places requires only the current coordinates when traveling. 
However, when planning a commute route to work, more additional information, such as the starting and ending points, as well as the preference, should be specified to execute this tool.
To better interact with such complicated tools, it is necessary to train to use the tool in many different scenarios.
Thus, the \textbf{second challenge} is knowing which tool is more complicated and how to improve the ability to use these tools.

In this paper, we propose the \textbf{\ours},  a tool-learning framework to train LLM to use complicated tools in real-world scenarios. \ours contains two main phases:
(1) To tackle the first challenge, we first propose a multi-stage learning method to teach the LLM to use various tools from an easy-to-difficult curriculum;
(2) We propose an {\full (\abb)} technique to dynamically construct the dataset to improve the ability to use the complicated tool.

Specifically, the multi-stage learning method involves three training stages: (1) warm-up training, (2) in-category training, and (3) cross-category training.
In the \textit{warm-up training} stage, we feed the model with the required minimal toolset and aim to teach the model to schedule and execute the tool correctly.
Next, in the \textit{in-category training} stage, we aim at teaching the model to learn to select the proper tools among related candidates.
Finally, we employ the \textit{cross-category training} stage, which trains the model in the real-world application setting, where the candidate toolset is constructed by a tool retriever that conducts semantics matching between the user query and tool demonstrations.
After being trained under our multi-stage training, an LLM becomes more straightforward and applied to real application scenarios.

Since the usage of some tools varies significantly in different scenarios, more extensive training should be conducted to fully master them.
Hence, we introduce the \full (\abb), to customize the tool-use training dataset iteratively, which includes two phases: instance generation and updates with introspective feedback.
In the instance generation phase, we start with a diverse toolset and an initial set of tool-use instance data.
Then the demonstration of tools is taken as prompts to ChatGPT to generate diverse queries and then answer these queries through compositional reasoning with various tools.
Since intricate tools require more training data for LLM to fully master, the pre-created dataset is out of sync with the up-to-date LLM.

Therefore, we take the introspection of the LLM for using tools as the feedback and use this feedback to guide the dataset update phase.
Specifically, in this phase, we aim to generate more tool-use instances related to the intricate tools that are usually misused by the current LLM.
Compared to previous works, \abb facilitates the LLM to master more intricate tools and prevents it from overfitting to a subset of simple tools.
To verify the effectiveness of \ours, we conduct extensive experiments on controlled and real-world settings using a large-scale tool-use dataset.
Experimental results show that our proposed \ours outperforms the tuning-free (\eg ChatGPT and Claude) and tuning-based baselines (\eg GPT4Tools) in terms of four aspects, which demonstrates the effectiveness of our tool-learning framework in the real-world application scenario.

To sum up, our contributions can be summarized as follows:
\begin{enumerate*}[label=(\roman*)]
\item We propose the \ours, a tool-learning framework, teaching the LLM to use complicated tools in real-world scenarios.
\item We propose a multi-stage learning method to improve the ability of multiple tool selection from a large-scale toolset.
\item We propose an iterative training strategy \abb to improve the performance of using intricate tools by dynamically updating the dataset according to the model introspection.
\item Experiments on both seen and unseen toolsets show that the \ours effectively accesses various tools and achieves comparable and even better performance to proprietary LLMs (\eg ChatGPT).
\end{enumerate*}

\section{Related Work}\label{sec:related}

\begin{figure*}[t]
        \centering
	\includegraphics[width=1\textwidth]{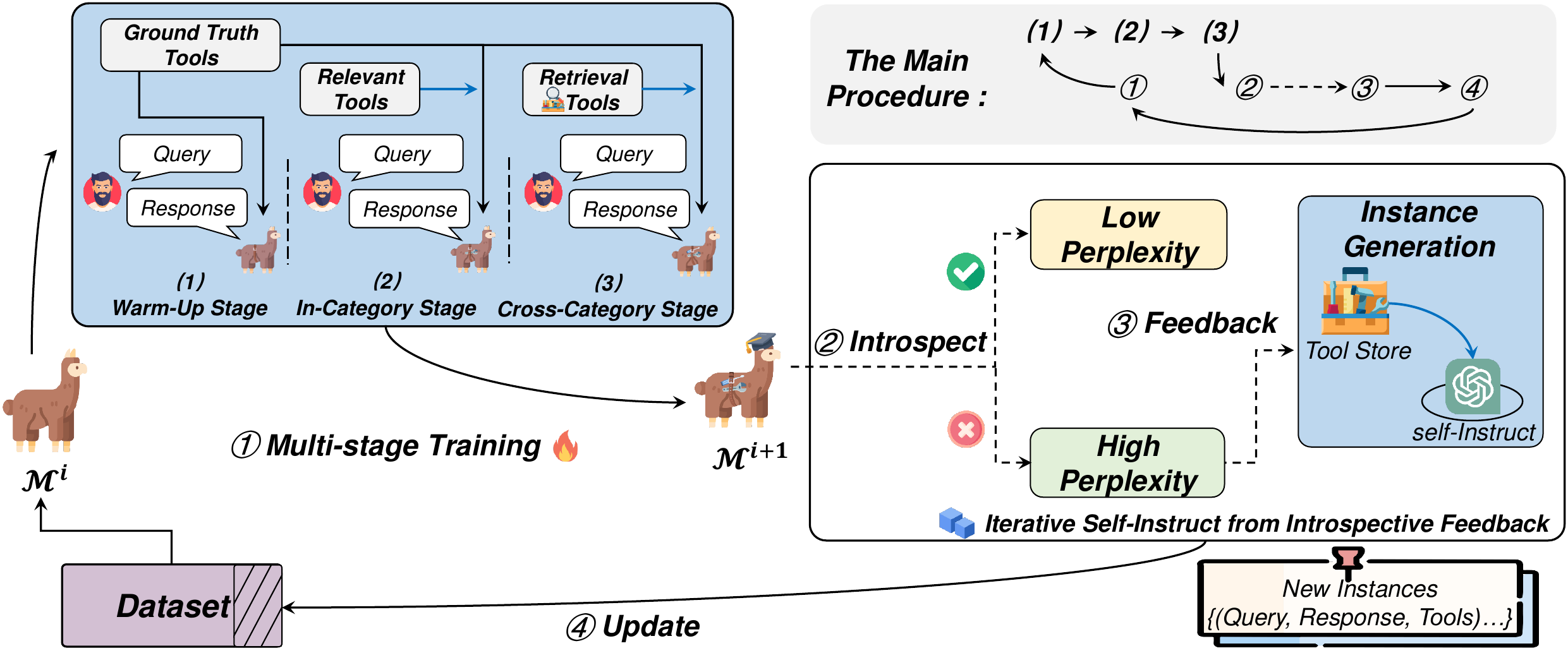}
        \caption{The overall architecture of our framework consists of multi-stage learning and iterative self-instruct from introspective feedback.
We denote the $\mathcal{M}^{i}$ as the target model trained on $i$-th epoch and the  $\mathcal{M}^{i+1}$ as the target model trained on $i+1$-th epoch.}
 \label{fig:overall}
\end{figure*}

\subsection{Tuning-free Tool Learning}

The tuning-free methods leverage the inherent in-context learning capability of LLMs, where the demonstrations of tools are taken as input to prompt LLMs to use various tools~\citep{art,react,rci}.
For example, \citeauthor{huggingfacegpt} and \citeauthor{visualgpt} integrate existing models hosted by Huggingface as the toolset to handle various downstream tasks, such as object detection and question answering.
The other studies, such as Chameleon~\citep{chameleon}, utilize the GPT-4  as the base model to devise long-term plans and automatically execute different tools, which further demonstrates the potential ability to tackle more complex tasks including table-based reasoning.
However, there are two main drawbacks of tuning-free methods:
(1) For data security reasons, not all the applications~\citep{Gao2019ProductAware} can transmit tool and user data to LLM service providers~\citep{gudibande2023false}. And it restricts the use of proprietary LLMs in such applications.
(2) Due to the limitation of the input length, the prompt cannot accommodate massive tools, thus constraining the model to utilize only a few tools to tackle the task.

\subsection{Tuning-based Tool Learning}

The tuning-based tool learning methods directly fine-tune the parameter of language models on the tool-use dataset~\citep{Wang2023DescribeEP}, typically constructed by prompting proprietary LLMs to use specific tools, \eg search~\citep{webcpm,webgpt}, calculation~\citep{toolkengpt,pal} and translation~\citep{toolformer}.
The advantage of these methods is that they can be easily deployed in a self-host environment.
However, fine-tuning langauge models on the constructed datasets typically introduces generalization problems~\citep{toolalpaca}, where performance degradation is usually observed when dealing with new tools which have not been seen during training.
To improve the generalization of tool-learning models for new tools, some works~\citep{toolw,xu2023tool, gorilla} devote to constructing datasets across diverse toolsets and increasing the diversity of training datasets, which present a promising solution to enhancing the performance of unseen tools.
However, they ignore the complexity distinctions between various tools which potentially leads to some complex tools with intricate usage are not well-learned, hurting the generalization of the model.
\section{Task Formulation}\label{sec:task}

We formulate the \ours as a tool learning framework to train an open-source large language model $\mathcal{M}$ to master various tools in real-world scenarios.
In detail, we start off a large toolset $\mathcal{T}^*$ with various tools and construct a tool-use dataset $D$.
Following \citep{api-bank}, we divide the tools into different categories (ten in our work), such as navigation and smart home.
Each instance $d$ in the dataset consists of the query ${q}=\{q_1,q_2,\dots,q_{|{q}|}\}$, response ${y}=\{y_1,y_2,\dots,y_{|{y}|}\}$ and ground truth tools $\mathcal{T}=\{\tau_1, \tau_2,\dots, \tau_{|\mathcal{T}|}\}$ for answering the query $q$.
Meanwhile, we denote the relevant toolset as $\mathcal{T}_\text{r}$, which derives from the same category as $\mathcal{T}$, and has no overlap with the $\mathcal{T}$.
We then train the target model $\mathcal{M}$ to decompose the original query $q$ into sub-tasks via compositional reasoning and schedule the appropriate tools step by step to generate the response $y$.
During inference, we first retrieve a subset $ \mathcal{\widetilde{T}}=\{\tau_1, \tau_2,\dots, \tau_{|\mathcal{\widetilde{T}}|}\}$ from toolset $\mathcal{T}^*$ for the given query ${q}$ which contains the candidate tools to generate the response.

Figure~\ref{fig:overall} shows the overall architecture of our proposed \ours operates the two main phrases iteratively:
(1) Given a tool-use dataset, we propose a multi-stage learning method to finetune the LLM in an easy-to-difficult curriculum paradigm;
(2) After tuning the LLM on the dataset, we dynamically update such dataset according to the confused set caused by the finetuned LLM. 
Continually, we employ the updated dataset to finetune the LLM and conduct the training paradigm in an iterative manner.

\section{Multi-stage Learning}\label{sec:toolllm}

In real-world applications, the tool-use model should select appropriate tools from the retrieved tools and schedule them correctly (\aka difficult mode), instead of directly using human given candidate toolset (\aka easy mode).
Similar to human learning procedures, tool learning models can benefit from an easy-to-difficult curriculum during model training~\citep{curriculum}. 
Therefore, we propose a multi-stage learning method that consists of warm-up training, in-category training, and cross-category training, teaching the LLM to master various tools in the real-world setup.

\subsection{Warm-Up Training Stage}\label{sec:warm-up}

In the initial warm-up stage, for each query $q$, we provide the LLM $\mathcal{M}$ with the ground truth toolset $\mathcal{T}$ to generate the response $y$, which can be formulated as:
\begin{equation}
   P(y | q, \mathcal{T})  = 
   \prod_{t=1}^{|y|} P_{\mathcal{M}}(y_t | y_{(<t)}, q, \mathcal{T}).
\end{equation}
Then we employ the log-likelihood objective $\mathcal{L}_{\text{warm-up}}$ to train the $\mathcal{M}$ to decompose the query into tool-use sub-tasks and generate the response $y$ by scheduling multiple tools:
\begin{equation}
   \mathcal{L}_{\text{warm-up}} = - \log P(y | q, \mathcal{T}).
   \label{eq:warm-up}
\end{equation}

\subsection{In-Category Training Stage}\label{sec:in-domain}

To gradually adapt the model to the real-world setting, for each query, we integrate a mixture of the ground truth toolset $\mathcal{T}$ and the relevant toolset $\mathcal{T}_\text{r}$ which are randomly selected from the same category as the $\mathcal{T}$.
The category of a tool indicates the using scenario, for example, planning a route and searching for a place are tools of the map navigation category.
In this setting, in addition to arranging the appropriate tools for the model, it is also necessary to first select the proper tools from candidates $\mathcal{T}_\text{r} \cup \mathcal{T}$.
And the LLM generates the response on the condition of the query $q$ and mixed toolset $\mathcal{T}_\text{r} \cup \mathcal{T}$, which can be formulated as:
\begin{equation}
   \mathcal{L}_{\text{in}} = - \sum_{t=1}^{|y|}  P_{\mathcal{M}}(y_t | y_{(<t)},q, \mathcal{T},\mathcal{T}_\text{r}).
   \label{eq:in}
\end{equation}

\subsection{Cross-Category Training Stage}\label{sec:hybrid}

Since the tools used to answer the query should be retrieved automatically rather than manually provided in real-world applications, we introduce the cross-category training method, which explicitly empowers LLM to select appropriate tools in the realistic setting.
Specifically, we first construct a tool retriever model based on the dual-encoder framework~\citep{sbert} to retrieve the candidate toolset $\mathcal{\widetilde{T}}$, which encodes the user query $q$ and the tool demonstrations into dense representations and computes the cosine similarity as relevance.  
Intuitively, the retrieved toolset $\mathcal{\widetilde{T}}$ contains the hard negative (redundant) tools that the LLM is more likely to get confused with. 
Therefore, we take the union of $\mathcal{T}$ and $\mathcal{\widetilde{T}}$ as the candidate toolset for each training example.
Then the LLM is supervised to select appropriate tools from the candidate toolset $\mathcal{\widetilde{T}} \cup \mathcal{T}$ and generate the response to query $q$, which can be formulated as:
\begin{equation}
   \mathcal{L}_{\text{cross}}=
   - \sum_{t=1}^{|y|}  P_{\mathcal{M}}(y_t | y_{(<t)},q, \mathcal{T},\mathcal{\widetilde{T}}).
   \label{eq:hybrid}
\end{equation}

\section{Iterative Self-Instruct from Introspection}
\label{sec:self-inst}

In order to conduct more targeted training for intricate tools, we propose the \textit{\full (\abb)}, a dynamic method for constructing training data, which updates the training dataset continuously based on model knowledge of tools.
As shown in Figure~\ref{fig:overall}, \abb iterates the two phases, \ie instance generation and  update with introspective feedback.

\subsection{Initial Dataset Construction}
\label{sec:init-dataset}

We start off building a tool store which contains 110 common-used tools and usage instances, which are constructed manually as the seed instance pool.
Specifically, each instance consists of a concrete query, and the answer follows the chain-of-thought format, where at least four tools are involved to encourage the complexity of our dataset.
As shown in Figure~\ref{fig:overall}, for each step, we first sample 5\textasciitilde7 tools from the tool store, denoted as $\mathcal{T}^*$.
Then, the demonstrations of sampled tools paired with corresponding instances are taken as input, prompting ChatGPT to reason the potential compositional relationship of tools and generate diverse instances.
In Table~\ref{tab:prompt}, we show an example of the prompt, which consists of three main parts: (1) task instruction; (2) candidate tools list; (3) tool-use instance demonstrations, which consist of a user query and a ground truth response.
More details for statistics and comparison with other related datasets are given in Table~\ref{tab:statistic}.
\begin{table}[t]
\tiny
\centering
\begin{tabular}{@{}m{8.3cm}@{}}
\toprule
\code{

\#\#\# You are an intelligent assistant with various tools.
You need to propose some real-world tasks and use the tools we provide to solve them.

\#\#\# You can use the following APIs:

SCHEDULE(string: n, time: t): 
schedule a meeting at time `t` on the topic of `n`.

...

EMAIL(user: x, string: s, string: c): send an email to the user `x` with the subject `s` and  content `c`. 

\#\#\# Here are some usage examples:

Query 1: Schedule a meeting for Michael about the new product launch. The meeting begins on June 1, 2023, 09:00:00. And send an e-mail for notification to...

Response 1: Schedule the meeting by [SCHEDULE(topic: New product launch, time: 2023-06-01 09:00:00)]. Then email to ...

Query 2: ...

Please come up with extra five queries and use the tools to solve them step-by-step.
Each query involves four tools at least.
}
\\
\bottomrule
\end{tabular}
\caption{Prompt used for generating new tool-use instances.}
\label{tab:prompt}
\end{table}
\begin{table}[t]
\small
\centering
\begin{tabular}{@{} l cccc @{}}
\toprule
\begin{tabular}[c]{@{}c@{}} Dataset \end{tabular}
& \begin{tabular}[c]{@{}c@{}}Tools \\  amount \end{tabular}
& \begin{tabular}[c]{@{}c@{}} Instance \\  amount \end{tabular}
& \begin{tabular}[c]{@{}c@{}} Reasoning \\ steps \end{tabular}  
&\begin{tabular}[c]{@{}c@{}} Avg. word \\  input/output  \end{tabular}
\\ 
\midrule
{API-bank } 
& 53
&  272
&   2.08    
&  56.50 / 59.39  
\\

{APIBench} 
& 3 
&  17,002    
&     1.0 
&      32.36 / 110.21 
\\

{ToolAlpaca}
& \textbf{426} 
&    3,938  
&   1.6
& 23.42 /  36.19     
\\

{Toolformer}
& 4
&  144,467 
&       -    
&   - 
 \\

{ToolBench}
& 8 
&  2,746   
& 5.9   
&   -     
\\

{\textbf{Ours}}
& 110 
&  \textbf{ 72,000 }  
&       \textbf{4.70}    
&         \textbf{223.66} / \textbf{75.59 }
\\

\bottomrule
\end{tabular}
\caption{Comparison of our and the other tool-use datasets.}
\label{tab:statistic}
\end{table}

\subsection{Updates with Introspective Feedback}\label{sec:ai-feeback}

Since the instances generated via self-instruct may be uncontrolled without any training targeted guidance~\citep{wizardlm,bian2023drop}, we propose to construct a prompt to guide the instance generation phase according to the training procedure.
Given a query containing $n$ tokens ${q}=\{q_1,... q_n\}$, we first retrieve a toolset $\mathcal{T^*}$, and then provide the LLM $\mathcal{M}$ with $\mathcal{T^*}$ to generate the response.
The generation perplexity $h$ of the target response which contains $m$ tokens ${y}=\{y_1,...y_m\}$ conditioned on ${q}$ and $\mathcal{T}^*$ can be factorized as follows:
\begin{equation}
    h = \sqrt[n]{\frac{1}{ P_{\mathcal{M}}({y} |{q},\mathcal{T}^*)}},
\end{equation}
where the $ P_{\mathcal{M}}({y} |{q},\mathcal{T}^*)$ is the generation probability, formulated as:
\begin{equation}
   P_{\mathcal{M}}({y} |{q},\mathcal{T}^*) = \prod_{i=1}^{|y|} P_{\mathcal{M}}(y_i|y_{(<i)}, {q},\mathcal{T}^*).
\end{equation}
Since perplexity $h$ represents the degree of generation uncertainty, samples with higher perplexity $h$ requires further training in subsequent training.

And next, we filter the generated instances $D=\{d_1,d_2,\dots,d_{|D|}\}$ with high perplexity instances $D^*$ which should be trained more.
These filtered instances $D^*$ are then utilized in the self-instruct prompt to generate more similar tool-use instances for further training.
The instance generate method is the same as the initial dataset construction (as shown in \S~\ref{sec:init-dataset}), only the tool-use demonstration in the prompt is replaced by the filtered instance $d^{*} \in D^*$.
Specifically, for each update, we generate $\sigma$ percent new instances of the original dataset, which is guided by the filtered instances, and we append these instances to the original dataset $D$.
The updated dataset will be used to train the model in the next epoch, and this process is conducted iteratively for each epoch.

\section{Experimental Setup}\label{sec:experiment}

\subsection{Dataset}

To verify the effectiveness of \ours, we employ two test sets: \textit{Seen} and \textit{Unseen} toolset, and each of them consists of 2,000 instances with ten tools.
All the tools in the \textit{Seen} toolset have been used in the training set, while the tools in the \textit{Unseen} toolset have not been used when training.

\subsection{Evaluation Metrics}

Following \citet{api-bank} and \citet{toolalpaca}, we evaluate from four aspects: tool selection,  parameter correctness, compositional reasoning, and interaction fluency.
\textbf{Tool Selection} evaluates the capability to select correct tools from the candidate toolset.
Since multiple tools are involved for each instance, we employ the listwise metric, which calculates the NDCG~\citep{ndcg} score between the tools in the generated response and the ground-truth response.
\textbf{Parameter Correctness} measures the correctness of the input parameter type for the tools, which validates whether the LLM response conforms to the schema of the tool's interface.
\textbf{Compositional Reasoning} first identifies the topological order of tools in generated and ground-truth response and calculates the ROUGE-L score of two sequences of tools.
\textbf{Interaction Fluency} employs the average of ROUGE-1, ROUGE-2, and ROUGE-L scores as the similarity between the generated and ground-truth responses, which indicates whether the model comprehends the output of tools and delivers fluent responses.
We also employ the human evaluation where three well-educated master students are invited to evaluate 50 randomly sampled cases with a three-scale score in the following two aspects:
(1) \textit{Executability}: whether the multiple tools are invoked in a correct logical order to generate the response
(2) \textit{Fluency}: whether the response generated via model is human-like and fluent~\citep{shi2023rade}.

\subsection{Baselines}

We  compare our \ours with tuning-based baselines, including \textit{ToolFormer}-6B~\citep{toolformer},  \textit{ToolLLaMA}-7B~\citep{toolw} and GPT4Tools~\citep{gpt4tools}.
We also compare with tuning-free methods, including the proprietary LLMs (e.g., ChatGPT and GPT-3) and open-source models, which interact with various tools by in-context learning.
For a fair comparison, all the tuning-based methods use the dataset as ours, and all the baselines are conducted on the candidate toolset, which is retrieved by our dense tool retrieve model (as shown in \S~\ref{sec:hybrid}).
We use the top-$10$ tools with the highest cosine similarity as the candidate toolset.

\subsection{Implementation Details}~\label{sec:implement}
In our work, we take the LLaMA-7B\footnote{\url{https://huggingface.co/huggyllama/llama-7B}} as our base model.  
We vary the percent $\sigma$ in $\{10, 15, 20, 25, 30\} $ and find that the $\sigma=20$ achieves the best performance.
We optimize the model using deepspeed ZeRO  strategy~\citep{deepspeed} with the learning rate of $5e^{-5}$ and the weight decay coefficient of 0.01.
The training of our model can be done within 20 hours with 4 NVIDIA A100-PCIE-80GB GPUs.

\section{Experimental Results}\label{sec:result}

\begin{table*}[!t]
\centering
\small
\setlength\tabcolsep{2pt}

\begin{tabular}{@{}
p{4.1cm}cccc  cccc@{}}

\toprule
& \multicolumn{4}{c}{\textbf{Seen Toolset}} 
& \multicolumn{4}{c}{\textbf{Unseen Toolset}} 
\\
\cmidrule(lr){2-5} \cmidrule(lr){6-9} 
\textbf{Method}
&   \begin{tabular}[c]{@{}c@{}} {Tool} \\ {Selection } \end{tabular}   
&   \begin{tabular}[c]{@{}c@{}} {Parameter} \\ {Correctness } \end{tabular}   
&   \begin{tabular}[c]{@{}c@{}} {Compositional} \\ {Reasoning } \end{tabular}   
&   \begin{tabular}[c]{@{}c@{}} {Interaction} \\ {Fluency } \end{tabular}   
&   \begin{tabular}[c]{@{}c@{}} {Tool} \\ {Selection } \end{tabular}   
&   \begin{tabular}[c]{@{}c@{}} {Parameter} \\ {Correctness } \end{tabular}   
&   \begin{tabular}[c]{@{}c@{}} {Compositional} \\ {Reasoning } \end{tabular}   
&   \begin{tabular}[c]{@{}c@{}} {Interaction} \\ {Fluency } \end{tabular}   
\\

\hline
\rowcolor{Gainsboro} \multicolumn{9}{l}{\textit{Tuning-free Methods}} \\
Claude
& 75.30	& 56.00	& 74.18 &	55.81
&45.13	&31.82&	45.71 &	51.38
\\

ChatGPT
&83.62	&67.31	&82.59	&65.65
&57.65	&48.24 & 57.05	& 56.89
\\

Text-davinci-003
& 79.13	& 59.71 & 	78.66	& 60.57
&54.73&	29.53 &54.72&46.37
\\

ChatGLM-6B~\citep{glm}
& 41.74 & 30.81  & 41.11 & 43.62
& 11.13	& 3.32	& 11.47& 	42.53
\\

ChatGLM2-6B~\citep{glm}
& 24.34	& 18.33	& 24.43	& 39.32
& 7.41 & 	5.62	& 7.75	& 22.37
\\

Llama-7B~\citep{gorilla}
&67.52	&53.81	&65.33	&47.71
&17.39	 &14.14	&18.20	&33.68
\\

Llama2-7B~\citep{gorilla}
& 70.93 & 	54.52 &	67.84	& 58.49
& 29.27	& 20.37	& 27.12& 	39.43
\\

Vicuna-7B~\citep{vicuna2023}
& 66.79	& 51.19	& 65.60	& 58.72
& 31.32	& 24.57	& 30.19	& 41.04
\\

Vicuna-13B~\citep{vicuna2023}
    & 72.26 & 	57.51	& 71.17	& 61.75
& 36.13	& 27.52	& 35.56	& 41.04
\\

\hline
\rowcolor{Gainsboro} \multicolumn{9}{l}{\textit{Tuning-based Methods}} \\

GPT4Tools~\citep{gpt4tools}
& 75.20	& 58.52	& 74.07  & 	64.99
& 44.58  &	30.21 	& 46.21 &		55.87 
\\

ToolLLaMA~\citep{toolw}
& 62.92	 & 44.92	&  62.99 & 	62.26 
& 34.33	 & 22.31  & 34.62	 &  50.62
\\

Toolformer~\citep{toolformer}
&  30.81 & 	 20.48  & 	29.65	& 38.75
&  29.27	&  22.36	& 27.16	& 36.29
\\

\midrule
\textbf{Ours (LLaMA-7B)} 
& \textbf{88.61	}&  \textbf{77.72} & 	\textbf{87.99} & 	\textbf{79.09}
& \textbf{59.79}	&  \textbf{50.21}	& \textbf{63.65} & 	\textbf{66.82}
\\

\midrule
\hline
\rowcolor{Gainsboro} \multicolumn{9}{l}{\textit{Ablation Study}} \\

\texttt{- w/o}  $ \mathcal{L}_{\text{warm-up}}$
& $86.11_{\downarrow2.50}$	& $70.16_{\downarrow7.56}$	& $83.91_{\downarrow4.08}$	&$ 74.86_{\downarrow4.23}$
& $57.30_{\downarrow2.49} $& 	$45.32_{\downarrow4.89}$ &  	$59.58_{\downarrow4.07}$ & 	$62.60_{\downarrow4.22} $
\\

\texttt{- w/o}     $\mathcal{L}_{\text{in}}$
& $85.73_{\downarrow2.88}$ &	$70.21_{\downarrow7.51}$ &	$83.64_{\downarrow4.35}$ &	$75.21_{\downarrow3.88}$ 
& $57.23_{\downarrow2.56}$ &	$47.22_{\downarrow2.99}$ &	$60.24_{\downarrow3.41}$ &	$62.27_{\downarrow4.55}$ 
\\

\texttt{- w/o} $\mathcal{L}_{\text{cross}}$
& $83.49_{\downarrow5.12}$ & 	$63.73_{\downarrow13.99}$ & 	$77.79_{\downarrow10.20}$	& $70.73_{\downarrow8.36}$
& $53.51_{\downarrow6.28} $ & 	$40.70_{\downarrow9.51} $& 	$53.46_{\downarrow10.19}$ &  	$58.46_{\downarrow8.36} $
\\

\texttt{- w/o} \abb 
& $83.52_{\downarrow5.09}$ & 	$ 67.46_{\downarrow10.26}$ & 	$81.21_{\downarrow6.78}$ &  	$73.05_{\downarrow6.04}$
& $54.72_{\downarrow5.07} $	& $41.45_{\downarrow8.76}$ 	& $56.87_{\downarrow6.78}$ & 	$60.71_{\downarrow6.11}$
\\

\midrule
\hline
\rowcolor{Gainsboro} \multicolumn{9}{l}{\textit{Effectiveness Analysis}} \\

Ours (LLaMA2-7B)
& 89.40  & 	77.81  & 84.38 & 75.22 	
& 59.84  & 	 49.53 & 	64.41 &   	68.82 
 
\\

Ours (Vicuna-7B)
& 87.30  & 	73.50   &  83.02  &  	76.04 
&56.94  & 48.62	  & 61.96 &  65.05
\\

\bottomrule
\end{tabular}
\caption{
Comparing with baselines on \textit{seen} and \textit{unseen} test datasets. The tools in the seen test set have been used in the training dataset, and the tools in the unseen test set have not been used when training. All the tuning-free methods learn to use the tools by in-context learning, and the tuning-based baselines use the same dataset as ours. 
}
\label{tab:main}

\end{table*}

\begin{table}[!t]
\centering
\small
\setlength\tabcolsep{3pt}

\begin{tabular}{@{}l cc  cc  @{}}

\toprule
& \multicolumn{2}{c}{\textbf{Seen Toolset}} 
& \multicolumn{2}{c}{\textbf{Unseen Toolset}} 
\\
\cmidrule(lr){2-3} \cmidrule(lr){4-5} 
\textbf{Method}
& Executability & Fluency
& Executability & Fluency
\\
\hline
\rowcolor{Gainsboro} \multicolumn{5}{l}{\textit{Tuning-free Methods}} \\
ChatGPT
& 2.70 & 2.73
& 2.07 & 2.04
\\

Text-divinci-003
& 2.53 & 2.50
& 1.85 & 2.15
\\
\hline
\rowcolor{Gainsboro} \multicolumn{5}{l}{\textit{Tuning-based Methods}} \\
Toolformer
& 1.19 & 1.22 
& 1.10 & 1.08
\\
GPT4tools
& 1.83 & 1.78
& 1.32 & 1.22
\\
\textbf{Ours}
& 2.78 & 2.80
& 2.00 & 2.14
\\
\bottomrule
\end{tabular}
\caption{Human evaluation on \textit{seen} and \textit{unseen} test datasets.}
\label{tab:human1}
\end{table}

\subsection{Overall Performance}

Table~\ref{tab:main} shows the experimental results of all baselines.
We can find that our proposed \ours achieves the best performance in seen and unseen toolsets in terms of all metrics.
Compared with ChatGPT, \ours gets 88.61 (4.99 absolute improvement) in terms of tool selection in the seen test set, which suggests \ours shows great potential for selecting proper tools correctly.
We observe that the \ours reaches 87.99 and 63.65 in the compositional reasoning aspect with the seen and unseen toolset, which has a significant improvement compared with the tuning-based baseline, and it also outperforms the advanced proprietary LLM, \ie ChatGPT.
This result highlights the \ours benefits from the chain-of-thought tool-use instances to conduct compositional reasoning.

In the unseen toolset, the \ours outperforms the strong tuning-free methods ChatGPT and Claude, pushing the score of tool selection to 59.79 ( 3.91\% relative improvement), demonstrating that \ours achieves effective generalized tool-use capability.
From Table~\ref{tab:main}, we  also find that the previous tuning-based  baselines suffer from a performance drop when generalizing from seen to unseen toolset.
For example, the compositional reasoning score of GPT4Tools is 74.07 with seen toolset while only 46.21 in the unseen toolset, which has a 37.61\% relative decrease.
The same trend has also been observed with ToolLLaMA (45.04\% relative decrease).
In contrast, \ours shows a slight decline, where the score of compositional reasoning has only a 27.65\% relative decrease.
The potential reason is that the LLM can acquire robust  tool-use skills from our iterative training strategy \abb.

Since the candidate toolset of all the baselines is retrieved automatically, we also verify the effectiveness of our tool retriever,
The \textit{recall@10} of our tool retriever achieves 93.49 and 91.20 in seen and unseen toolsets.
It shows that the tool retriever with dual-encoder architecture is qualified to find proper tools closely aligned with the ground truth.

\subsection{Human Evaluation}

We conduct human evaluation, and Table~\ref{tab:human1} summarizes the results.
We find that the \ours consistently outperforms the best tuning-based baselines in two aspects, such as pushing \textit{Executability} to 2.73 (0.90 absolute improvement) with seen toolset.
Moreover, we also observe that the \ours achieves comparable or even better results with ChatGPT, indicating the effectiveness of our framework.
The average Kappa statistics for two evaluation metrics are 0.762 and 0.732, illustrating agreement among the annotators.
We also provide examples of model outputs in the supplementary material.

\begin{figure}[t]
        \centering
	\includegraphics[width=0.48\textwidth]{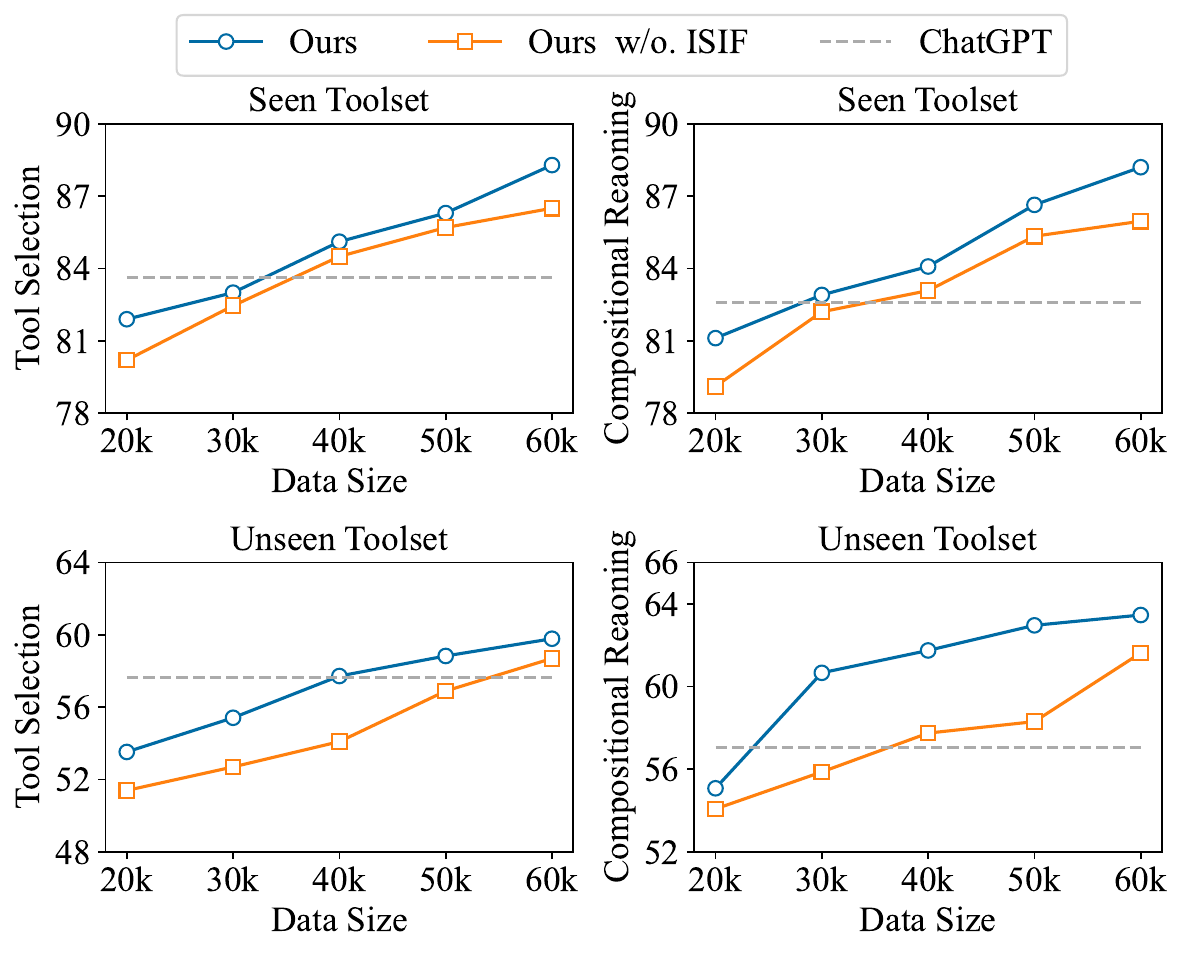}
        \caption{Comparison between \ours with \abb and a variant model which randomly sample tools to generate new instances without the introspective feedback.}
 \label{fig:size-performace}
\end{figure}

\subsection{Analysis of Multi-stage Training}

In Table~\ref{tab:main}, we compare \ours with several ablation variants, including the model (w/o $\mathcal{L}_{\text{warm-up}}$, $\mathcal{L}_{\text{in}}$, and $\mathcal{L}_{\text{cross}}$) which removes each training stage in the multi-stage training method.
We can find that all the variant models suffer performance degradation, which demonstrates the effectiveness of our proposed multi-stage training methods in \ours.
We observe that the model w/o $\mathcal{L}_{\text{cross}}$ has the largest performance drop compared with the other two variant models in terms of the tool selection score.
This phenomenon demonstrates the necessity of constructing a candidate toolset similar to the real-world setting to improve the tool selection ability of LLM.

\subsection{Analysis of \abb}

We explore whether the performance improvement is simply caused by the expansion of the training set so as to further verify the necessity of the introspective feedback in \abb.
For a fair comparison, different from \abb, which updates the dataset according to the high perplexity instance, we random sample some instances as the prompt of self-instruct to generate new instances. 
And then, the updated dataset is used to train the LLaMA, which is the same base model as our \ours.
Figure~\ref{fig:size-performace} shows the performance of the models trained on different sizes of initial datasets.
We find that our proposed \abb performs constantly better than the model directly trained by vanilla self-instruct in each size of the dataset, which verifies the effectiveness of dynamically updating the dataset guided by the introspective feedback.

\begin{figure}[t]
        \centering
	\includegraphics[width=0.48\textwidth]{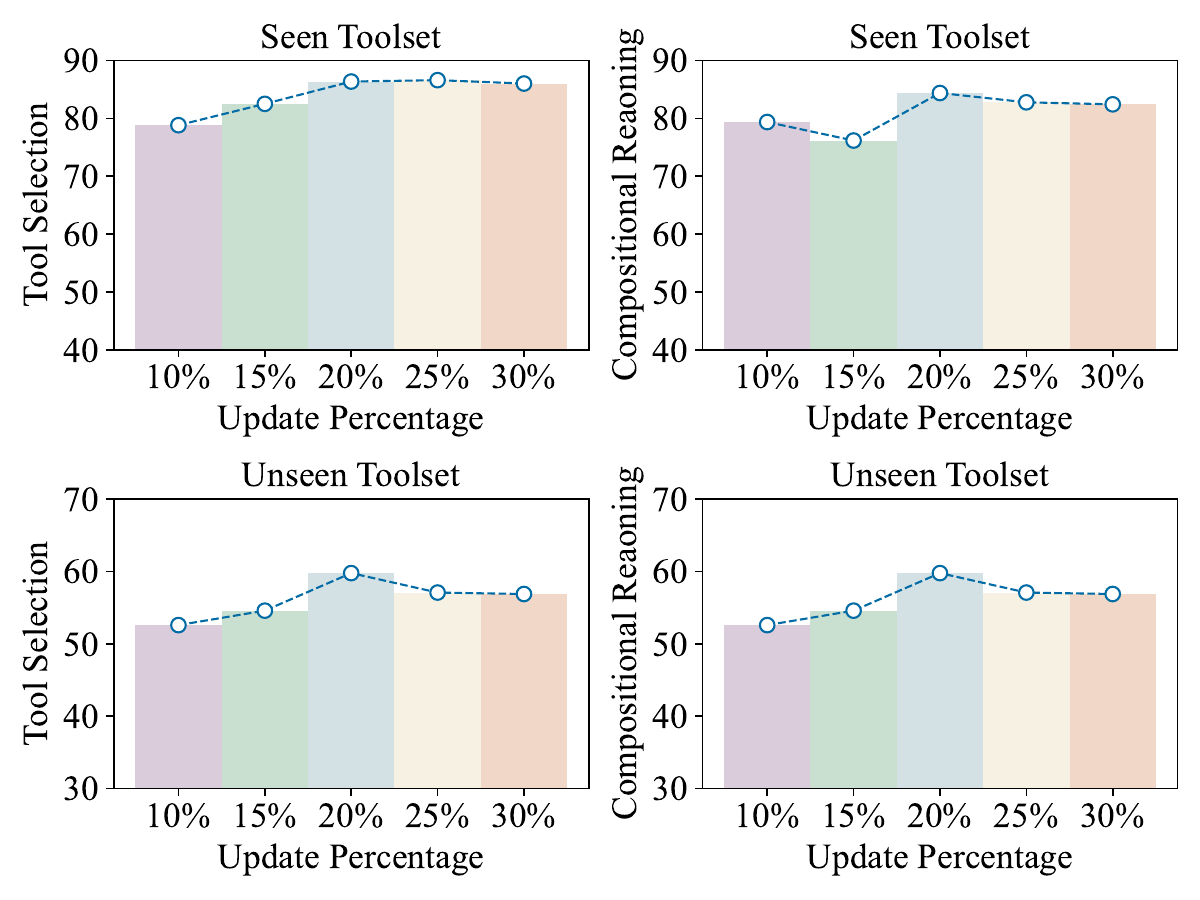}
        \caption{The qualitative analysis for update percentage.}
 \label{fig:update-performace}
\end{figure}

\subsection{Generalization for Different Base Models}

To further explore the robustness of our proposed \ours, we finetune the other two open-source LLMs (LLaMA2-7B and Vicuna-7B) using \ours with the same setting shown in \S~\ref{sec:implement}.
As Table~\ref{tab:main} shows, compared with the corresponding tuning-free versions, both two models trained using \ours outperform their base model by a large margin, demonstrating the generalization of our framework. 

\subsection{Qualitative Analysis for Update Percentage}
In our \ours, we dynamically update the $\sigma$ percent dataset according to the perplexity.
These filtered instances are used as the self-instruct prompt to generate new tool-use instances (as shown in \S~\ref{sec:ai-feeback}).
By appending these new instances to the training set, the model can enhance the understanding of these tools.
In this section, we explore the effect of the percentage of data updates $\sigma$ on the final performance.
In Figure~\ref{fig:update-performace}, we vary the data updating percentage $\sigma$ from 10\% to 30\%. 
As the percentage $\sigma$ changes from 10\% to 20\%, the performance keeps increasing and peaks at 20\%.
It can be seen from the results that with the increase of the dynamic update percentage, the performance is also improved. This phenomenon can prove the effectiveness of our proposed updating with introspective feedback. 
But as the update percentage continues to increase to 30\%, the performance begins to decline. 
One possible reason is that too many targeted tool-use instances introduce the distribution bias, which causes the model to overfit a few specific tools, thereby reducing the generalization of the model.

\section{Conclusion}\label{sec:conclusion}

In this paper, we propose the \ours, a novel tool learning framework to teach LLM to master various tools, which consists of two main steps: (1) multi-stage learning and (2) iterative self-instruct from introspective feedback (\abb).
Concretely, we fine-tune the LLM with three learning stages from an easy-to-difficult curriculum, \ie warm-up, in-category, and cross-category stages.
Since the usage of some tools varies in different scenarios, which requires more training to fully understand the usage, we introduce the ISIF to iteratively update the tool-use training dataset based on the model introspection.
Extensive experiments on seen and unseen toolsets demonstrate that \ours can boost the tool-learning performance of LLM compared with both tuning-based and tuning-free baselines, including ChatGPT.

\section*{Acknowledgements}\label{sec:ack}
This work was supported by the National Natural Science Foundation of China (T2293773).

\bibliography{aaai24}

\begin{thebibliography}{32}
\providecommand{\natexlab}[1]{#1}

\bibitem[{Bian et~al.(2023)Bian, Liu, Han, Lin, Lu, He, and Sun}]{bian2023drop}
Bian, N.; Liu, P.; Han, X.; Lin, H.; Lu, Y.; He, B.; and Sun, L. 2023.
\newblock A Drop of Ink Makes a Million Think: The Spread of False Information
  in Large Language Models.
\newblock \emph{ArXiv preprint}, abs/2305.04812.

\bibitem[{Chiang et~al.(2023)Chiang, Li, Lin, Sheng, Wu, Zhang, Zheng, Zhuang,
  Zhuang, Gonzalez, Stoica, and Xing}]{vicuna2023}
Chiang, W.-L.; Li, Z.; Lin, Z.; Sheng, Y.; Wu, Z.; Zhang, H.; Zheng, L.;
  Zhuang, S.; Zhuang, Y.; Gonzalez, J.~E.; Stoica, I.; and Xing, E.~P. 2023.
\newblock Vicuna: An Open-Source Chatbot Impressing GPT-4 with 90\%* ChatGPT
  Quality.

\bibitem[{Du et~al.(2022)Du, Qian, Liu, Ding, Qiu, Yang, and Tang}]{glm}
Du, Z.; Qian, Y.; Liu, X.; Ding, M.; Qiu, J.; Yang, Z.; and Tang, J. 2022.
\newblock {GLM}: General Language Model Pretraining with Autoregressive Blank
  Infilling.
\newblock In \emph{Proceedings of the 60th Annual Meeting of the Association
  for Computational Linguistics (Volume 1: Long Papers)}, 320--335. Dublin,
  Ireland: Association for Computational Linguistics.

\bibitem[{Gao et~al.(2023)Gao, Madaan, Zhou, Alon, Liu, Yang, Callan, and
  Neubig}]{pal}
Gao, L.; Madaan, A.; Zhou, S.; Alon, U.; Liu, P.; Yang, Y.; Callan, J.; and
  Neubig, G. 2023.
\newblock {PAL}: Program-aided Language Models.
\newblock In \emph{PMLR}, 10764--10799.

\bibitem[{Gao et~al.(2019)Gao, Ren, Zhao, Zhao, Yin, and
  Yan}]{Gao2019ProductAware}
Gao, S.; Ren, Z.; Zhao, Y.; Zhao, D.; Yin, D.; and Yan, R. 2019.
\newblock Product-Aware Answer Generation in E-Commerce Question-Answering.
\newblock In \emph{WSDM}.

\bibitem[{Gudibande et~al.(2023)Gudibande, Wallace, Snell, Geng, Liu, Abbeel,
  Levine, and Song}]{gudibande2023false}
Gudibande, A.; Wallace, E.; Snell, C.~B.; Geng, X.; Liu, H.; Abbeel, P.;
  Levine, S.; and Song, D. 2023.
\newblock The False Promise of Imitating Proprietary LLMs.
\newblock \emph{ArXiv preprint}, abs/2305.15717.

\bibitem[{Hao et~al.(2023)Hao, Liu, Wang, and Hu}]{toolkengpt}
Hao, S.; Liu, T.; Wang, Z.; and Hu, Z. 2023.
\newblock ToolkenGPT: Augmenting Frozen Language Models with Massive Tools via
  Tool Embeddings.
\newblock \emph{ArXiv preprint}, abs/2305.11554.

\bibitem[{J\"{a}rvelin and Kek\"{a}l\"{a}inen(2002)}]{ndcg}
J\"{a}rvelin, K.; and Kek\"{a}l\"{a}inen, J. 2002.
\newblock Cumulated Gain-Based Evaluation of IR Techniques.
\newblock \emph{TOIS}, 422–446.

\bibitem[{Jin et~al.(2023)Jin, Yang, Chen, and Lu}]{Jin2023GeneGPTAL}
Jin, Q.; Yang, Y.; Chen, Q.; and Lu, Z. 2023.
\newblock GeneGPT: Augmenting Large Language Models with Domain Tools for
  Improved Access to Biomedical Information.
\newblock \emph{ArXiv}.

\bibitem[{Kim, Baldi, and McAleer(2023)}]{rci}
Kim, G.; Baldi, P.; and McAleer, S. 2023.
\newblock Language Models can Solve Computer Tasks.
\newblock \emph{ArXiv preprint}, abs/2303.17491.

\bibitem[{Li et~al.(2023)Li, Song, Yu, Yu, Li, Huang, and Li}]{api-bank}
Li, M.; Song, F.; Yu, B.; Yu, H.; Li, Z.; Huang, F.; and Li, Y. 2023.
\newblock API-Bank: A Benchmark for Tool-Augmented LLMs.
\newblock \emph{ArXiv preprint}, abs/2304.08244.

\bibitem[{Lu et~al.(2023)Lu, Peng, Cheng, Galley, Chang, Wu, Zhu, and
  Gao}]{chameleon}
Lu, P.; Peng, B.; Cheng, H.; Galley, M.; Chang, K.-W.; Wu, Y.~N.; Zhu, S.-C.;
  and Gao, J. 2023.
\newblock Chameleon: Plug-and-Play Compositional Reasoning with Large Language
  Models.
\newblock \emph{ArXiv preprint}, abs/2304.09842.

\bibitem[{Nakano et~al.(2021)Nakano, Hilton, Balaji, Wu, Ouyang, Kim, Hesse,
  Jain, Kosaraju, Saunders, Jiang, Cobbe, Eloundou, Krueger, Button, Knight,
  Chess, and Schulman}]{webgpt}
Nakano, R.; Hilton, J.; Balaji, S.~A.; Wu, J.; Ouyang, L.; Kim, C.; Hesse, C.;
  Jain, S.; Kosaraju, V.; Saunders, W.; Jiang, X.; Cobbe, K.; Eloundou, T.;
  Krueger, G.; Button, K.; Knight, M.; Chess, B.; and Schulman, J. 2021.
\newblock WebGPT: Browser-assisted question-answering with human feedback.
\newblock \emph{ArXiv preprint}, abs/2112.09332.

\bibitem[{Paranjape et~al.(2023)Paranjape, Lundberg, Singh, Hajishirzi,
  Zettlemoyer, and Ribeiro}]{art}
Paranjape, B.; Lundberg, S.~M.; Singh, S.; Hajishirzi, H.; Zettlemoyer, L.; and
  Ribeiro, M.~T. 2023.
\newblock ART: Automatic multi-step reasoning and tool-use for large language
  models.
\newblock \emph{ArXiv preprint}, abs/2303.09014.

\bibitem[{Park et~al.(2023)Park, O'Brien, Cai, Morris, Liang, and
  Bernstein}]{Park2023GenerativeAI}
Park, J.~S.; O'Brien, J.~C.; Cai, C.~J.; Morris, M.~R.; Liang, P.; and
  Bernstein, M.~S. 2023.
\newblock Generative Agents: Interactive Simulacra of Human Behavior.
\newblock \emph{ArXiv preprint}, abs/2304.03442.

\bibitem[{Patil et~al.(2023)Patil, Zhang, Wang, and Gonzalez}]{gorilla}
Patil, S.~G.; Zhang, T.; Wang, X.; and Gonzalez, J.~E. 2023.
\newblock Gorilla: Large Language Model Connected with Massive APIs.
\newblock \emph{ArXiv preprint}, abs/2305.15334.

\bibitem[{Qin et~al.(2023{\natexlab{a}})Qin, Cai, Jin, Yan, Liang, Zhu, Lin,
  Han, Ding, Wang, Xie, Qi, Liu, Sun, and Zhou}]{webcpm}
Qin, Y.; Cai, Z.; Jin, D.; Yan, L.; Liang, S.; Zhu, K.; Lin, Y.; Han, X.; Ding,
  N.; Wang, H.; Xie, R.; Qi, F.; Liu, Z.; Sun, M.; and Zhou, J.
  2023{\natexlab{a}}.
\newblock {W}eb{CPM}: Interactive Web Search for {C}hinese Long-form Question
  Answering.
\newblock In \emph{ACL}, 8968--8988.

\bibitem[{Qin et~al.(2023{\natexlab{b}})Qin, Hu, Lin, Chen, Ding, Cui, Zeng,
  Huang, Xiao, Han, Fung, Su, Wang, Qian, Tian, Zhu, Liang, Shen, Xu, Zhang,
  Ye, Li, Tang, Yi, Zhu, Dai, Yan, Cong, Lu, Zhao, Huang, Yan, Han, Sun, Li,
  Phang, Yang, Wu, Ji, Liu, and Sun}]{toolw}
Qin, Y.; Hu, S.; Lin, Y.; Chen, W.; Ding, N.; Cui, G.; Zeng, Z.; Huang, Y.;
  Xiao, C.; Han, C.; Fung, Y.~R.; Su, Y.; Wang, H.; Qian, C.; Tian, R.; Zhu,
  K.; Liang, S.; Shen, X.; Xu, B.; Zhang, Z.; Ye, Y.; Li, B.; Tang, Z.; Yi, J.;
  Zhu, Y.; Dai, Z.; Yan, L.; Cong, X.; Lu, Y.-T.; Zhao, W.; Huang, Y.; Yan,
  J.-H.; Han, X.; Sun, X.; Li, D.; Phang, J.; Yang, C.; Wu, T.; Ji, H.; Liu,
  Z.; and Sun, M. 2023{\natexlab{b}}.
\newblock Tool Learning with Foundation Models.
\newblock \emph{ArXiv preprint}, abs/2304.08354.

\bibitem[{Rasley et~al.(2020)Rasley, Rajbhandari, Ruwase, and He}]{deepspeed}
Rasley, J.; Rajbhandari, S.; Ruwase, O.; and He, Y. 2020.
\newblock DeepSpeed: System Optimizations Enable Training Deep Learning Models
  with Over 100 Billion Parameters.
\newblock In Gupta, R.; Liu, Y.; Tang, J.; and Prakash, B.~A., eds.,
  \emph{{KDD} '20: The 26th {ACM} {SIGKDD} Conference on Knowledge Discovery
  and Data Mining, Virtual Event, CA, USA, August 23-27, 2020}, 3505--3506.
  {ACM}.

\bibitem[{Reimers and Gurevych(2019)}]{sbert}
Reimers, N.; and Gurevych, I. 2019.
\newblock Sentence-{BERT}: Sentence Embeddings using {S}iamese {BERT}-Networks.
\newblock In \emph{Proceedings of the 2019 Conference on Empirical Methods in
  Natural Language Processing and the 9th International Joint Conference on
  Natural Language Processing (EMNLP-IJCNLP)}, 3982--3992. Hong Kong, China:
  Association for Computational Linguistics.

\bibitem[{Schick et~al.(2023)Schick, Dwivedi-Yu, Dess{\`i}, Raileanu, Lomeli,
  Zettlemoyer, Cancedda, and Scialom}]{toolformer}
Schick, T.; Dwivedi-Yu, J.; Dess{\`i}, R.; Raileanu, R.; Lomeli, M.;
  Zettlemoyer, L.; Cancedda, N.; and Scialom, T. 2023.
\newblock Toolformer: Language Models Can Teach Themselves to Use Tools.
\newblock \emph{ArXiv preprint}, abs/2302.04761.

\bibitem[{Shen et~al.(2023)Shen, Song, Tan, Li, Lu, and
  Zhuang}]{huggingfacegpt}
Shen, Y.; Song, K.; Tan, X.; Li, D.~S.; Lu, W.; and Zhuang, Y.~T. 2023.
\newblock HuggingGPT: Solving AI Tasks with ChatGPT and its Friends in
  HuggingFace.
\newblock \emph{ArXiv preprint}, abs/2303.17580.

\bibitem[{Shi et~al.(2023)Shi, Sun, Zhang, Zhang, Ren, and Ren}]{shi2023rade}
Shi, Z.; Sun, W.; Zhang, S.; Zhang, Z.; Ren, P.; and Ren, Z. 2023.
\newblock RADE: Reference-Assisted Dialogue Evaluation for Open-Domain
  Dialogue.
\newblock \emph{arXiv preprint arXiv:2309.08156}.

\bibitem[{Tang et~al.(2023)Tang, Deng, Lin, Han, Liang, and Sun}]{toolalpaca}
Tang, Q.; Deng, Z.; Lin, H.; Han, X.; Liang, Q.; and Sun, L. 2023.
\newblock ToolAlpaca: Generalized Tool Learning for Language Models with 3000
  Simulated Cases.
\newblock \emph{ArXiv preprint}, abs/2306.05301.

\bibitem[{Wang et~al.(2023)Wang, Cai, Liu, Ma, and Liang}]{Wang2023DescribeEP}
Wang, Z.; Cai, S.; Liu, A.; Ma, X.; and Liang, Y. 2023.
\newblock Describe, Explain, Plan and Select: Interactive Planning with Large
  Language Models Enables Open-World Multi-Task Agents.
\newblock \emph{ArXiv}, abs/2302.01560.

\bibitem[{Wu et~al.(2023)Wu, Yin, Qi, Wang, Tang, and Duan}]{visualgpt}
Wu, C.; Yin, S.-K.; Qi, W.; Wang, X.; Tang, Z.; and Duan, N. 2023.
\newblock Visual ChatGPT: Talking, Drawing and Editing with Visual Foundation
  Models.
\newblock \emph{ArXiv preprint}, abs/2303.04671.

\bibitem[{Xu et~al.(2020)Xu, Zhang, Mao, Wang, Xie, and Zhang}]{curriculum}
Xu, B.; Zhang, L.; Mao, Z.; Wang, Q.; Xie, H.; and Zhang, Y. 2020.
\newblock Curriculum Learning for Natural Language Understanding.
\newblock In \emph{Proceedings of the 58th Annual Meeting of the Association
  for Computational Linguistics}, 6095--6104. Online: Association for
  Computational Linguistics.

\bibitem[{Xu et~al.(2023{\natexlab{a}})Xu, Sun, Zheng, Geng, Zhao, Feng, Tao,
  and Jiang}]{wizardlm}
Xu, C.; Sun, Q.; Zheng, K.; Geng, X.; Zhao, P.; Feng, J.; Tao, C.; and Jiang,
  D. 2023{\natexlab{a}}.
\newblock WizardLM: Empowering Large Language Models to Follow Complex
  Instructions.
\newblock \emph{ArXiv preprint}, abs/2304.12244.

\bibitem[{Xu et~al.(2023{\natexlab{b}})Xu, Hong, Li, Hu, Chen, and
  Zhang}]{xu2023tool}
Xu, Q.; Hong, F.; Li, B.; Hu, C.; Chen, Z.; and Zhang, J. 2023{\natexlab{b}}.
\newblock On the Tool Manipulation Capability of Open-source Large Language
  Models.
\newblock \emph{ArXiv preprint}, abs/2305.16504.

\bibitem[{Yang et~al.(2023{\natexlab{a}})Yang, Song, Li, Zhao, Ge, Li, and
  Shan}]{gpt4tools}
Yang, R.; Song, L.; Li, Y.; Zhao, S.; Ge, Y.; Li, X.; and Shan, Y.
  2023{\natexlab{a}}.
\newblock GPT4Tools: Teaching Large Language Model to Use Tools via
  Self-instruction.
\newblock \emph{ArXiv preprint}, abs/2305.18752.

\bibitem[{Yang et~al.(2023{\natexlab{b}})Yang, Li, Wang, Lin, Azarnasab, Ahmed,
  Liu, Liu, Zeng, and Wang}]{yang2023mm}
Yang, Z.; Li, L.; Wang, J.; Lin, K.; Azarnasab, E.; Ahmed, F.; Liu, Z.; Liu,
  C.; Zeng, M.; and Wang, L. 2023{\natexlab{b}}.
\newblock MM-REACT: Prompting ChatGPT for Multimodal Reasoning and Action.
\newblock \emph{ArXiv preprint}, abs/2303.11381.

\bibitem[{Yao et~al.(2023)Yao, Zhao, Yu, Du, Shafran, Narasimhan, and
  Cao}]{react}
Yao, S.; Zhao, J.; Yu, D.; Du, N.; Shafran, I.; Narasimhan, K.~R.; and Cao, Y.
  2023.
\newblock ReAct: Synergizing Reasoning and Acting in Language Models.
\newblock In \emph{ICLR}.

\end{thebibliography}
\clearpage
\appendix

\section{Appendix} \label{app}

\subsection{Data Quality Review}\label{app:data-filter}
To encourage the diversity of our dataset, a new instance is added to the instance pool only when its ROUGE-L~\footnote{\url{https://torchmetrics.readthedocs.io/en/stable/}} similarity with any existing instance is less than 0.7 to encourage diversity.
And we filter the instances with less than four required tools to encourage the complexity of the dataset.
The invalid instances are also excluded based on heuristics rules (\eg input parse errors, input type errors, and invalid tool usages).

To investigate the quality of our dataset, we invite three well-educated students to review random sample 1000 instances in the following aspects:
\begin{enumerate*}[label=(\roman*)]
\item reasonableness of generated query;
\item tool selection; 
\item tool-use normativity; 
\item response fluency 
\end{enumerate*}
Specifically, we provide each annotator the review guideline shown in Table~\ref{tab:quality}, where each question in the guideline is annotated on a binary scale (1 for yes and 0 for no).
We can observe that most of the generated instances are meaningful (more than 90\%), while the generated instances may contain noise (within 10\%).
We further analyze the generated instance, which is annotated with invalid.
And we find that even though the generations may contain errors, most of them are still in the correct logic, which can provide useful guidance for training models in the tool learning task.
he average Kappa statistics for the human evaluation is 0.891, illustrating agreement amount the annotators.

\begin{table}[htbp]
\small
\centering
\setlength{\tabcolsep}{2mm}{
\begin{tabular}{@{}p{6cm} c@{}}
\toprule
\begin{tabular}[c]{@{}c@{}} \textbf{Quality Review Question} \end{tabular}    
&\begin{tabular}[c]{@{}c@{}} Yes or No \% \end{tabular}    
\\
\cmidrule(r){1-1}
\cmidrule(r){2-2}

\textit{whether the generated  question exhibit a reasonable disposition and necessitate the integration of the external toosl.}
& \multirow{3}{*}{ 90.1\% / 9.9\%}     
\\

\specialrule{0em}{1pt}{1pt}
\cdashline{1-2}[6pt/6pt]
\specialrule{0em}{1pt}{1pt}

\textit{Whether the model select the appreciate tools and generate correct response to the query.}
 & \multirow{2}{*}{ 92.3\% / 8.7\%}    
\\

\specialrule{0em}{1pt}{1pt}
\cdashline{1-2}[6pt/6pt]
\specialrule{0em}{1pt}{1pt}

\textit{Does the model adhere to the prescribed conventions for invoking external tools, encompassing aspects such as function nomenclature, parameter configuration, and other pertinent formatting criteria?} 
&  \multirow{5}{*}{ 89.2\% / 10.8\%}   
\\

\specialrule{0em}{1pt}{1pt}
\cdashline{1-2}[6pt/6pt]
\specialrule{0em}{1pt}{1pt}

\textit{Whether the generated response is fluent and human-like without observe grammatical issues.}
& 
 \multirow{3}{*}{ 94.1\% / 5.9\%}  
\\

\bottomrule
\end{tabular}
}

\caption{The  quality review for our initial dataset, where the four questions are reviewed by three well-educated masters with 0.891 \textit{Kappa} statistics.
}
\label{tab:quality}
\end{table}

\subsection{Data collection}\label{app:data-collection}

We propose the \textit{\abb} to customize the dataset for different base models, where the initial dataset is updated during the training process base on the current knowledge of the model for the tools.
We also release the initial dataset to facilitate future research.
Table~\ref{tab:app-statistic} summarizes the statistic of our dataset and comparison with the existing related datasets.
\begin{table}[htbp]
\small
\centering
\begin{tabular}{@{} l cccc @{}}
\toprule
\begin{tabular}[c]{@{}c@{}} Dataset \end{tabular}
& \begin{tabular}[c]{@{}c@{}}Tools \\  amount \end{tabular}
& \begin{tabular}[c]{@{}c@{}} Instance \\  amount \end{tabular}
& \begin{tabular}[c]{@{}c@{}} Reasoning \\ steps \end{tabular}  
&\begin{tabular}[c]{@{}c@{}} Avg. word \\  input/output  \end{tabular}
\\ 
\midrule
{API-bank } 
& 53
&  272
&   2.08    
&  56.50 / 59.39  
\\

{APIBench} 
& 3 
&  17,002    
&     1.0 
&      32.36 / 110.21 
\\

{ToolAlpaca}
& \textbf{426} 
&    3,938  
&   1.6
& 23.42 /  36.19     
\\

{Toolformer}
& 4
&  144,467 
&       -    
&   - 
 \\

{ToolBench}
& 8 
&  2,746   
& 5.9   
&   -     
\\

{\textbf{Ours}}
& 110 
&  \textbf{ 72,000 }  
&       \textbf{4.70}    
&         \textbf{223.66} / \textbf{75.59 }
 \\

\bottomrule
\end{tabular}
\caption{Comparison of our and the other tool-use datasets.}
\label{tab:app-statistic}
\end{table}

\begin{figure}[htbp]
        \centering
	\includegraphics[width=0.5\textwidth]{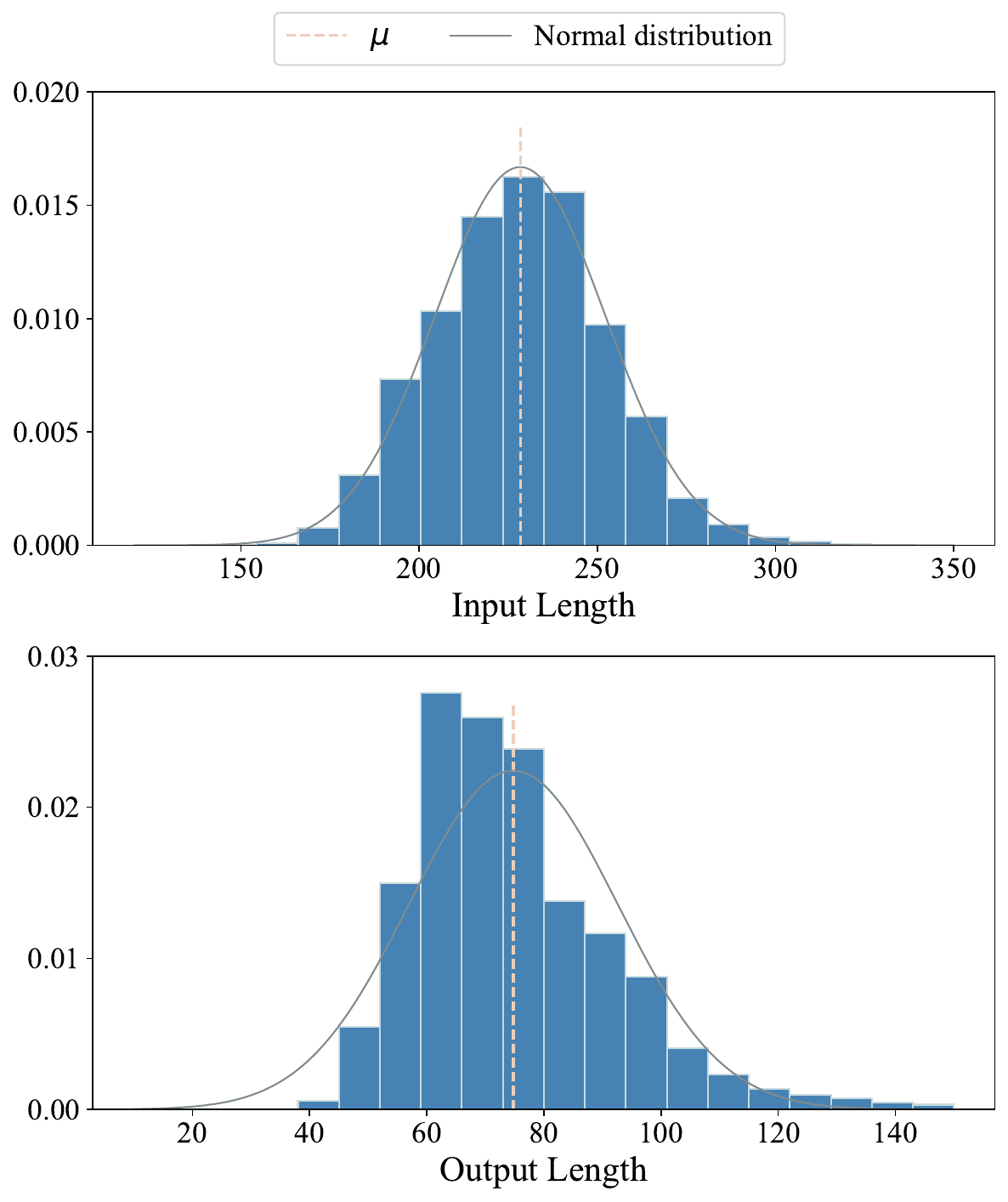}
        \caption{The input and output length distribution of our initial dataset.}
 \label{fig:length}
\end{figure}

\begin{figure}[htbp]
        \centering
	\includegraphics[width=0.5\textwidth]{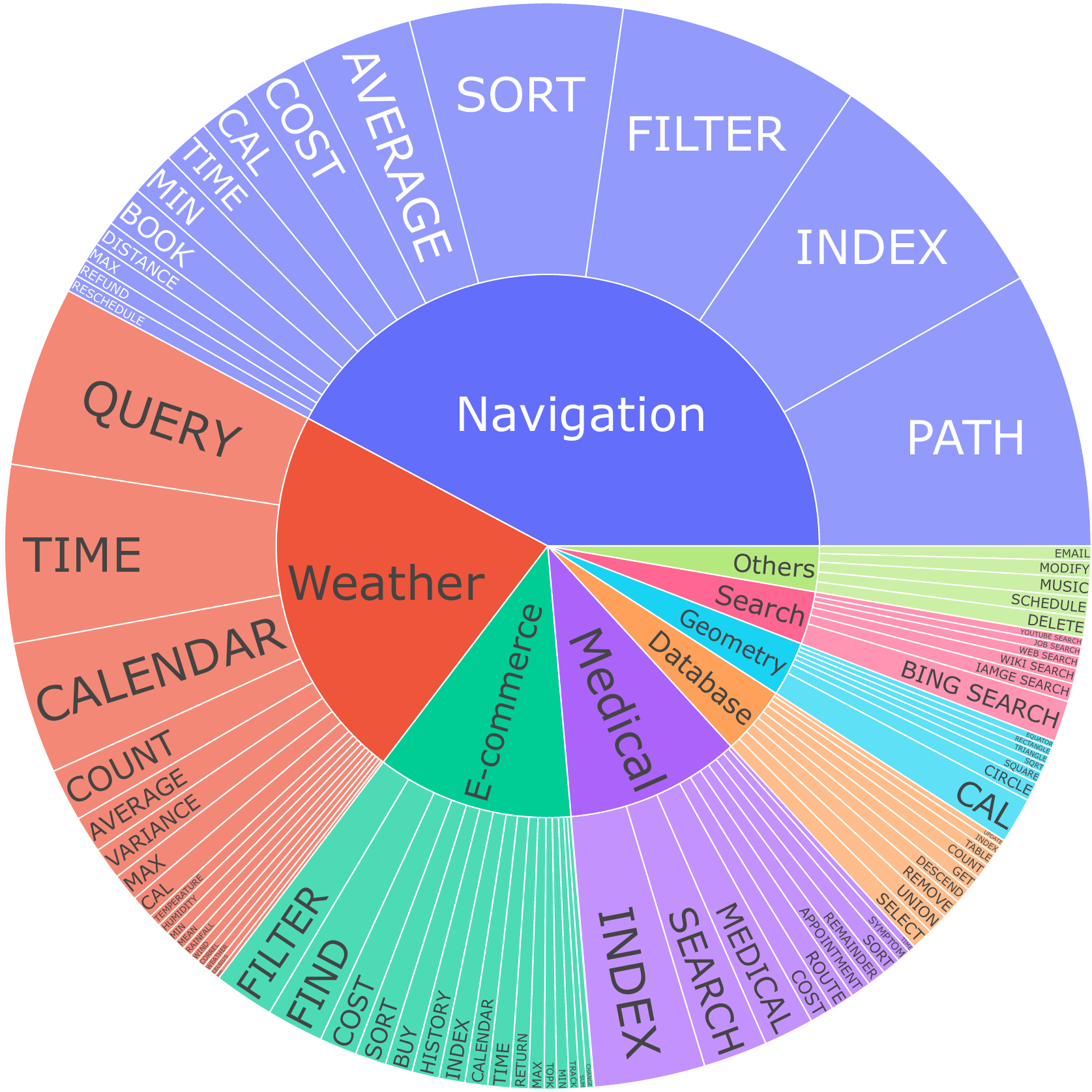}
        \caption{The most common-used tools in our tool store and corresponding category. }
 \label{fig:pie}
\end{figure}

And we  demonstrate diversity in the length of the instance inputs, and instance outputs in Figure~\ref{fig:length}.
To explain the diversity of our dataset more intuitively, we plot~\footnote{\url{https://github.com/matplotlib/matplotlib}} the top 20 most common tools in our tool store in Figure~\ref{fig:pie}.


\subsection{Evaluation Details}
For the tuning-based methods in our experiments, we consider the zero-shot setting as the out-of-the-box configuration where only API documentation is provided without demonstration examples. We use this configuration to understand the initial gap in capabilities among models.
And all the tuning-based methods are fine-tuned in our dataset with the same setting to ensure the fairness of the comparison.

 \subsection{Case study}\label{app:case}
 
 We conduct several case studies and find that our method is more effective at incorporating external tools and generating more fluent responses than baselines.
Table~\ref{tab:case1}, Table~\ref{tab:case2}  and Table~\ref{tab:case3} present examples of the response generated by our method and strong baselines.
Since the execution result (\aka the return value) o externals tools are in the complex \textit{json} format, we represent the same \textbf{execution result} with the same \textbf{placeholder} (e.g., \%s).

 \begin{table*}[htbp]
\small \centering
\setlength\tabcolsep{4pt}
\begin{tabular}{@{}rp{14cm}@{}}
\toprule
\textbf{Tool category} &{Navigation}\\
\midrule
\multirow{1}{*}{\textbf{Query}}
& Can you help me find the average distance and average travel time from Berlin to Munich by train? 
\\
\midrule
\multirow{7}{*}{\textbf{Tool demonstration}}
& PATH(string: place1, string: place2) → list: get the list of paths from place1 to place2. The place1 and place2 are the string: type. 
\\
& SORT(list: paths, string: f) → list: sort the list of path `paths` based on the feature `f`, such as the cost, distance, and time. And the `f` is the string type.
\\
& FILTER(list: paths, string: f) → list: filter the list of path `paths` based on feature `f` (string type) and keep only the paths that match the feature (e.g., cost, distance, and time).
\\
& AVERAGE(list: paths, string: f) → path: get the average of the path list `paths` based on the feature `f` (string type, e.g., cost, distance, and time).
\\
& MIN(list: paths, string: f)  → path: get the path from the path list `paths` with the minimum feature.
\\
& MAX(list: paths, string: f) → path: get the path from the path list `paths` with the maximum feature.
\\
& GET\_COST(path: p) → cost: get the cost of the path `p`.
\\
& GET\_DISTANCE(string: place1, string: place2): get the distance from  `place1` to  `place2`, which are the string type.
\\
& GET\_TIEM(path: p) → time: get the total time of the path `p`.
\\
& INDEX(list: paths, int: i) → path: get the i-th path from the list of the path `paths`.
\\
& CALCULATE(expression: e) → float: calculate the expression `e` result, e.g., 1+2, 1/3, 4*5, and 7-1. The expression `e` can be about time, money, or distance
\\
& BOOK(path: p) → ticket: book a ticket of the path `p`.
\\
& REFUND(ticket: t) → money: return the ticket `t` and get the money.
\\
& RESCHEDULE(ticket: t, path: p) → ticket: reschedule the ticket to a new path `p` and get the new ticket.
\\

\hline
\rowcolor{Gainsboro} \multicolumn{2}{l}{\textbf{ \xspace\xspace\xspace\xspace\xspace\xspace    Model outputs}}\\
\textit{Ours} &
The route from Berlin to Munich is \textcolor{teal}{[PATH(string: Berlin, string: Munich) → \%s1]}. Among these routes, \textcolor{teal}{[FILTER(list: \%s1, string: train) → \%s2]} needs to be taken by train. The average distance is \textcolor{teal}{[AVERAGE(list: \%s2, string: distance) → \%s3]}. The average travel time is \textcolor{teal}{[AVERAGE(list: \%s2, string: time) → \%s4]}. \#\#\# The average distance from Berlin to Munich by train is \%s3 and the average travel time is \%s4.\\

\specialrule{0em}{1pt}{1pt}
\cdashline{1-2}[6pt/6pt]
\specialrule{0em}{1pt}{1pt}

\textit{ChatGPT} &
The route from Berlin to Munich by train is \textcolor{teal}{[PATH(string: Berlin, string: Munich) → \%s1]}. The average distance can be calculated by finding the average of the distance feature for the routes in \%s1. To do this, sort the routes according to distance as \textcolor{red}{[SORT(list: \%s1, string: distance) → \%s2]}. Then, find the average distance by using the AVERAGE function as  \textcolor{teal}{[AVERAGE(list: \%s2, string: distance) → \%s3]}. Similarly, the average travel time can be obtained by sorting the routes in \%s1 according to travel time as \textcolor{red}{[SORT(list: \%s1, string: time) → \%s4]}, and then finding the average travel time using the AVERAGE function as \textcolor{teal}{[AVERAGE(list: \%s4, string: time) → \%s5]}. \#\#\# The average distance from Berlin to Munich by train is \%s3 and the average travel time is \%s4. \\

\specialrule{0em}{1pt}{1pt}
\cdashline{1-2}[6pt/6pt]
\specialrule{0em}{1pt}{1pt}

\textit{GPT4Tools} &
The route from Berlin to Munich is \textcolor{teal}{[PATH(string: Berlin, string: Munich) → \%s1]}. Among these routes, \textcolor{teal}{[FILTER(list: \%s1, string: train) → \%s2]} needs to be taken the train. Sort these train routes according to distance as \textcolor{red}{[SORT(list: \%s2, string: distance) → \%s3]}. The average distance is \textcolor{teal}{[AVERAGE(list: \%s3, string: distance) → \%s4]}. The average travel time is \textcolor{red}{[AVERAGE(list: [FILTER(list: \%s1, string: train) → \%s2], string: time) → \%s5]}. \#\#\# The average distance from Berlin to Munich by train is \%s4, and the average travel time is \%s5.
\\

\midrule
\textit{Ground Truth }&
The route from Berlin to Munich is \textcolor{teal}{[PATH(place: Berlin, place: Munich) → \%s1]}. The train routes available are \textcolor{teal}{[FILTER(list: \%s1, string: train) → \%s2]}. The average distance from Berlin to Munich by train is \textcolor{teal}{[AVERAGE(list: \%s2, string: distance) → \%s3]}, and the average travel time is \textcolor{teal}{[AVERAGE(list: \%s2, string: time) → \%s4]}. \#\#\# The average distance from Berlin to Munich by train is \%s3, and the average travel time is \%s4.
\\
\bottomrule
\end{tabular}
\caption{
An example of model outputs on the seen toolset (\textit{Navigation} tools), where we highlight the \textcolor{red}{wrong} and \textcolor{teal}{right} tool-use.
This example shows that the tuning-free methods which rely on the advanced proprietary LLM, e.g., ChatGPT,  encounter difficulty selecting appropriate tools from the candidate toolset in realistic scenarios.
In contrast, our model selects the correct tools and generate fluent response for the query, indicating the effectiveness of our multi-stage learning method. We represent the same execution result with the same placeholder (e.g., \%s).  }
\label{tab:case1}
\end{table*}

\begin{table*}[htbp]
\small \centering
\setlength\tabcolsep{4pt}
\begin{tabular}{@{}rp{14cm}@{}}
\toprule
\textbf{Tool category} &{Weather System}\\
\midrule
\multirow{1}{*}{\textbf{Query}}
&  What is the mean rainfall in London during the past month?
\\
\midrule
\multirow{7}{*}{\textbf{Tool demonstration}}
& WEATHER(city: c, string: d) →  weather: get the weather of the city `c` on the date `d`. And the `d` is the string with ``yyyymmdd`` format to indicate the date. \\
& TEMPERATURE(city: c, string: d) → temperature: get the temperature of the city `c` in the date of `d`. And the `d` is the string with ``yyyymmdd`` format to indicate the date. \\
& HUMIDITY(city: c, string: d) → humidity: get the humidity of the city `c` in the date of  `d`. And the `d` is the string with ``yyyymmdd`` format to indicate the date. \\
& WIND(city: c, string: d) → speed: get the wind speed of the city `c` in the date of `d`. And the `d` is the string with ``yyyymmdd`` format to indicate the date. \\
& AVERAGE(list: l) → value: get the average  of the list `l` from the date `d1` to date `d2`. The feature `f` indicates the temperature, humidity, etc.\\
& MIN(list: l) → value: get the minimum of the list `l` from the date `d1` to date `d2`. The feature `f` indicates the temperature, humidity, etc.\\
& MAX(list: l) → value: get the maximum  of the list `l` from the date `d1` to date `d2`. The feature `f` indicates the temperature, humidity, etc.\\
& VARIANCE(list: l, string: d1, string: d2) → value: get the variance of the city `c` from the date `d1` to date `d2`. The feature `f` indicates the temperature, humidity, etc.\\
& COUNT(list: paths, string: f): count the number of occurrences of feature `f` (string type) in list `l`.\\
& RAINFALL(city: c, date: y) → rainfall: get the rainfall of the city `c` in the year `y`. And the `y` is the string with ``yyyymmdd`` format to indicate the date. \\
& CELSIUS(fahrenheit: f) → celsius: convert Fahrenheit `f` to celsius.\\
& FAHRENHEIT(celsius: c) → fahrenheit: convert celsius `c` to fahrenheit.\\
& CAL(expression: e)->float: calculate the result of expression `e`, e.g., 1+2, 1/3, 4*5 and 7-1.\\
& CALENDAR() → date:  get the date of today.\\
& TIME(string: d, int: n): get the time of `n' days before date `d` (d<0) or after date `d` (d>0). And the `d1` and `d2` are the string with ``yyyymmdd`` format to indicate the date. \\
& QUERY(city: c, string: d1, date: d2, string: f) → list: get the list of feature of city `c` from the date `d1` to date `d2`. And the `d1` and `d2` are the string with ``yyyymmdd`` format to indicate the date. \\

\hline
\rowcolor{Gainsboro} \multicolumn{2}{l}{\textbf{ \xspace\xspace\xspace\xspace\xspace\xspace    Model outputs}}\\
\textit{Ours} &
Today is \first{[CALENDAR() -> \%s1]}, and the past month is \second{[TIME(date: \%s1, int: -30) -> \%s2]}. The rainfall in London during the past month is \third{[RAINFALL(city: London, date: \%s2) -> \%s3]}. The mean rainfall in London during the past month is \fouth{[MEAN(list: \%s3) -> \%s4]}. \#\#\# The mean rainfall in London during the past month is \%s4.
\\

\specialrule{0em}{1pt}{1pt}
\cdashline{1-2}[6pt/6pt]
\specialrule{0em}{1pt}{1pt}

\textit{ChatGPT} &
 Today is \first{[CALENDAR() -> \%s1]}, and one month ago is \second{[TIME(date: \%s1, int: -30) -> \%s2]}. The rainfall in London from \%s2 to \%s1 is \third{[QUERY(city: London, date: \%s2, date: \%s1, string: rainfall) -> \%s3]}. And the mean rainfall is \fouth{[MEAN(list: \%s3) -> \%s4]}. \#\#\# The mean rainfall in London during the past month is \%s4.\\

\specialrule{0em}{1pt}{1pt}
\cdashline{1-2}[6pt/6pt]
\specialrule{0em}{1pt}{1pt}

\textit{GPT4Tools} &
\third{[QUERY(city: London, date: 2022-04-01, date: 2022-05-01, string: rainfall) -> list]}
\\

\midrule
\textit{Ground Truth }&
Today is \first{[CALENDAR() -> \%s1]}, and one month ago is \second{[TIME(date: \%s1, int: -30) -> \%s2]}. The rainfall in London from \%s2 to \%s1 is \third{[QUERY(city: London, date: \%s2, date: \%s1, string: rainfall) -> \%s3]}. And the mean rainfall is \fouth{[MEAN(list: \%s3) -> \%s4]}. \#\#\# The mean rainfall in London during the past month is \%s4.\\

\bottomrule
\end{tabular}
\caption{
An example of model outputs on the unseen toolset (\textit{Weather System} tools), where the tools remain unseen during the training process.
The same tools are highlighted in the same color for a more intuitive explanation.
We find that our model selects appropriate tools and generates the desired response as the advanced proprietary LLM, i.e. ChatGPT, though generalizing to the unseen toolsets.
These results demonstrate that  LLM can acquire robust tool-use skills from our iterative training strategy \abb. The placeholders (e.g., \%s) indicates the execution results of our tools.
}
\label{tab:case2}
\end{table*}

\begin{table*}[htbp]
\small \centering
\setlength\tabcolsep{4pt}
\begin{tabular}{@{}rp{14cm}@{}}
\toprule
\textbf{Tool category} &{Medical Care}\\
\midrule
\multirow{1}{*}{\textbf{Query}}
& Please find a suitable doctor for Peter as he is experiencing chest pain.
\\
\midrule
\multirow{7}{*}{\textbf{Tool demonstration}}
& SEARCH(string: s) → list: search for a list of disease names based on the given symptom `s`.
\\
& MEDICAL(disease: d) → list: get the list of medicine to cure the disease `d`.
\\
& ROUTE(string: s) → doctor: match a suitable doctor according to the symptoms `s`.
\\
& APPOINTMENT(user: u, doctor: t) → date: make an appointment with doctor `t`.
\\
& SYMPTOM (disease: d) → string: search the symptom of the given disease `d`.
\\
& COST(medical: m) → : get the cost of the medical `m`.
\\
& REMAINDER(date: d, record: r) → bool: set a record `r` on the memo on a date `d`.
\\
& CALENDAR() → date:  get the date of today.
\\
& TIME(date: d, int: n) → date: get the time of `n' days before date `d` (d<0) or after date `d` (d>0).
\\
& SORT(list: l, string: f) → list: sort the list `l` based on the feature `f`.
\\
& INDEX(list: l, int: i) → item: get the i-th item of the list `l`.
\\
\hline
\rowcolor{Gainsboro} \multicolumn{2}{l}{\textbf{ \xspace\xspace\xspace\xspace\xspace\xspace    Model outputs}}\\
\textit{Ours} &
Chest pain can be caused by various conditions, for example \third{[SEARCH(string: chest pain) →\%s1]}. Dr. \second{[ROUTE(string: chest pain) →\%s2]} specializes in treating chest pain. You can make an appointment with this doctor by \first{[APPOINTMENT(user: Peter, doctor:\%s2) →\%s3]}. And this appointment has been added to your reminder via \fouth{[REMAINDER(date:\%s3, record: appointment with\%s2) →\%s4]}. \#\#\# Appointment with\%s2 doctor to\%s3 for consultation.
\\

\specialrule{0em}{1pt}{1pt}
\cdashline{1-2}[6pt/6pt]
\specialrule{0em}{1pt}{1pt}

\textit{ChatGPT} &
Chest pain can be caused by various conditions, for example \third{[SEARCH(string: chest pain) →\%s1]}. Dr. \second{[ROUTE(string: chest pain) →\%s2]} is a suitable doctor for his symptoms.
\\

\specialrule{0em}{1pt}{1pt}
\cdashline{1-2}[6pt/6pt]
\specialrule{0em}{1pt}{1pt}

\textit{GPT4Tools} &
Dr. \second{[ROUTE(string: chest pain) →\%s1] } is a suitable doctor for Peter's symptoms. You can make an appointment with this doctor via \first{[APPOINTMENT(user: Peter, doctor:\%s1) →\%s2]}. And this appointment has been added to your remainder via \fouth{[REMAINDER(date:\%s2, record: appointment with\%s1) →\%s3]}. \#\#\# Appointment with Dr.\%s1 for consultation.
\\

\midrule
\textit{Ground Truth }&
Chest pain can be caused by various conditions, for example \third{[SEARCH(string: chest pain) →\%s1]}. Dr. \second{[ROUTE(string: chest pain) →\%s2]} specializes in chest-related conditions. An appointment with this doctor has been scheduled for \first{[APPOINTMENT(user: Peter, doctor:\%s2) →\%s3]}. And this appointment has been added to your reminder via \fouth{[REMAINDER(date:\%s3, record: appointment with\%s2) →\%s4]}. \#\#\# Appointment with\%s2 doctor to\%s3 for consultation.
\\
\bottomrule
\end{tabular}
\caption{
A complex example of the seen test dataset (\textit{Medical Care} tools), is where the model should reason the relationship between multiple tools and use the tools in a logical order.
We highlight the same tool-calling with the same color.
We can observe that the ChatGPT and GPT4Tools fail to use tools correctly, while our agents generate a desired response that answers the query.
This comparison provides a intuitive evidence for the effectiveness of the proposed tool-learning framework. We represent the same execution result with the same placeholder (e.g., \%s).
}
\label{tab:case3}
\end{table*}

\clearpage

\end{document}